\documentclass[11pt]{article}
\usepackage{amsmath}\begin{normalsize}
\end{normalsize}
\usepackage{algorithm}
\usepackage[noend]{algpseudocode}
\usepackage{color}
\usepackage{bm,verbatim}
\usepackage[authoryear]{natbib}
\usepackage[colorlinks,citecolor=blue,linkcolor=blue,urlcolor=blue]{hyperref}
\usepackage{cleveref}
\usepackage{latexsym, epsfig, amssymb, amsmath, amsthm, graphicx, mathrsfs, booktabs}
\usepackage{rotating, tabularx, setspace,paralist}
\usepackage{color}
\usepackage{graphicx}
\usepackage[OT1]{fontenc}
\usepackage{lscape}
\usepackage{multirow,multicol,mathtools}
\usepackage{subcaption}
\usepackage{anysize}
\renewcommand{\baselinestretch}{1.2}
\marginsize{1.1in}{1.1in}{0.55in}{0.8in} %
\newcommand{\boldres}[1]{{\textbf{\textcolor{red}{#1}}}}
\newcommand{\secondres}[1]{{\underline{\textcolor{blue}{#1}}}}

\usepackage{ifthen,amsmath,amssymb}

\makeatletter
\def\singlespace{\def\baselinestretch{1}\@normalsize}

\makeatother

\def\independenT#1#2{\mathrel{\setbox0\hbox{$#1#2$}%
\copy0\kern-\wd0\mkern4mu\box0}}

\def\beginn{\begin{eqnarray*}}
\def\endn{\end{eqnarray*}}
\def\beginy{\begin{eqnarray}}
\def\endy{\end{eqnarray}}
\def\begine{\begin{enumerate}}
\def\ende{\end{enumerate}}
\definecolor{bittersweet}{rgb}{0.8, 0.5, 0.2}

\newtheorem{assumption}{Assumption}

\newtheorem{theorem}{Theorem}

\newtheorem{example}{Example}

\newcommand{\bw}{\mbox{\bf w}}
\newcommand{\bx}{\mbox{\bf x}}

\newcommand{\var}{\mathrm{var}}

\definecolor{mygrey}{gray}{0.6} 

\allowdisplaybreaks

\usepackage[draft,inline,nomargin,index]{fixme}
\fxsetup{theme=color,mode=multiuser}
\FXRegisterAuthor{yf}{ayf}{\color{magenta} YF}
\renewcommand{\top}{T}

\renewcommand{\ldots}{\cdots}

\begin{document}

\title{PEARL: Performance-Enhanced Aggregated Representation Learning

\author{Wenhui Li$^{1}$\thanks{Wenhui Li and Shijing Gong contributed equally to this work.} , 
Shijing Gong$^{2}$\footnotemark[1], and 
Xinyu Zhang$^{1,2}$\thanks{Corresponding author: xinyu@amss.ac.cn}
\medskip\\
Academy of Mathematics and Systems Science, Chinese Academy of Sciences$^1$; \\ School of Management, University of Science and Technology of China$^2$
\\
} %
}

\maketitle

\begin{abstract} 
Representation learning is a key technique in modern machine learning that enables models to identify meaningful patterns in complex data. However, different methods tend to extract distinct aspects of the data, and relying on a single approach may overlook important insights relevant to downstream tasks. This paper proposes a performance-enhanced aggregated representation learning method, which combines multiple representation learning approaches to improve the performance of downstream tasks. The  framework is designed to be general and flexible, accommodating a wide range of loss functions commonly used in machine learning models. To ensure computational efficiency, we use surrogate loss functions to facilitate practical weight estimation. Theoretically, we prove that our method asymptotically achieves optimal performance in downstream tasks, meaning that the risk of our predictor is asymptotically equivalent to the theoretical minimum. Additionally, we derive that our method asymptotically assigns nonzero weights to correctly specified models. We evaluate our method on diverse tasks by comparing it with advanced machine learning models. The experimental results demonstrate that our method consistently outperforms baseline methods, showing its effectiveness and broad applicability in real-world machine learning scenarios.
\end{abstract}

\textit{Running title}: PEARL


\newpage
\section{Introduction}\label{Sec.new1}
Modern data analysis frequently involves high-dimensional, large-scale, structured, unstructured, and multi-modal datasets. Representation learning plays a crucial role in extracting information-rich features from such data, and it has become a fundamental technique across various machine learning domains, including natural language processing \citep{Vaswani2017NIPS}, computer vision \citep{NIPS2012imagenet}, and speech recognition \citep{radford2023robust}, and has also shown success in bioinformatics \citep{Alexander2021pnas} and neuroscience \citep{willem2023pnas}. Traditional representation learning methods, including factor analysis and multi-dimensional scaling, have been extended to deep learning-based approaches \citep{Xie2020ARSA}.
Deep learning-based approaches employ multi-layer nonlinear transformations to progressively capture abstract and hierarchical features from raw data. For instance, early neural network models demonstrated success in visual pattern recognition \citep{lecun1998gradient,NIPS2012imagenet}. More recent transformer-based architectures \citep{Vaswani2017NIPS} form the basis of modern large language models capable of learning representations and processing natural language text \citep{NEURIPS2020language, ouyang2022training}.

The performance of machine learning methods heavily relies on the choice of data representations \citep{bengio2013representation}. With the large pool of available representation learning methods, some studies have treated representation choice as a model selection problem, aiming to pick the most effective pretrained representation for a given task \citep{nguyen2020leep,You2021LogME}. However, relying on a single representation learning method may ignore complementary information that other methods may capture. Moreover, the effectiveness of a selected representation learning model can vary substantially across data and task conditions \citep{Ericsson2021IEEE, Agostinelli2022ECCV}. To address these issues, it is essential to develop strategies that aggregate multiple representation learning methods.

In this paper, we propose a model averaging framework called performance-enhanced aggregated representation learning (PEARL), designed to improve downstream performance by combining diverse representations. 
Model averaging is a popular technique that addresses the uncertainty of choosing models \citep{Hjort2003jasa, hansen2007eco}, and recently works have proven success in complex models including random forest \citep{chen2024optimal} and neural networks \citep{zeng2023collapsed,wang2025credal}. However, existing efforts are often confined to specific model structures or task domains. This raises the need for a more general model averaging framework that can accommodate both simple models and deep complex architectures, while being applicable across diverse tasks. Representation learning provides a natural bridge toward this goal since representation learning is a core component of most machine learning methods.

PEARL consists of two stages. In the first stage, a set of foundation representation learning (FRL) models, which are either trained from scratch or adapted from pretrained encoders such as BERT \citep{devlin2019bert} and GPT \citep{NEURIPS2020language}, transform raw inputs into representation vectors. These representations, individually or combined through integration strategies, form a pool of candidate representation sets. In the second stage, each candidate set is linked to the target task through a downstream predictor (e.g., a linear model or neural network).  The final output of PEARL is obtained by averaging these predictions, with weights estimated by a cross-validation-based weighting criterion.

The term ``performance-enhanced'' in PEARL reflects its effectiveness in both computational feasibility and theoretical guarantees. For the computational simplicity,  a key design principle of PEARL is the weighting procedure. To accommodate diverse downstream tasks, the weighting criterion is formulated on the basis of a generalized loss function. However, directly optimizing certain loss functions may introduce non-smooth or non-convex optimization challenges. To enable efficient and stable weight computation, PEARL employs surrogate loss functions as approximations to original loss functions. Moreover, PEARL adopts a linear weighting scheme rather than nonlinear deep aggregation, thereby further reducing computational cost and preserving interpretability.

We establish two main theoretical guarantees for PEARL. First, from the perspective of downstream task performance, we prove the asymptotic optimality of the PEARL predictor. Specifically, the predictive risk of PEARL converges to the lowest possible risk defined by the original loss. This result ensures that PEARL performs at least as well as any single FRL in the pool and other linear combinations of FRLs, such as simple ensemble. Second, from the perspective of weight interpretability, we prove weight consistency. Specifically, when the candidate pool contains correctly specified models, PEARL asymptotically assigns all weight to them. This property demonstrates that PEARL is able to identify and prioritize well-specified models among candidates.
Together, these results provide a solid theoretical foundation for both the effectiveness and the interpretability of PEARL across diverse downstream tasks.

The rest of the paper is organized as follows. Section \ref{Method} presents the model framework including representation learning process, downstream task analysis, and weight choice criterion. Section \ref{TheoreticalResults} shows the asymptotic optimality and weight consistency of the proposed method PEARL. Section \ref{Experiments} shows one numerical simulation study and four real-world datasets. Section \ref{Summary} summarizes our work and discusses the future works. In supplementary materials,  Appendix A  gives the proofs of theoretical results, and Appendix  B shows the implementation details of experiments. 

\section{Method}\label{Method}
Suppose that there is a set of different artificial intelligence techniques that map between the predictors $x \in \mathcal{X}$ and the outputs $y \in \mathcal{Y}$. We aim to predict an output in a downstream task given a new predictor.  
The predictor space $\mathcal{X}$ is generalized to accommodate a broad spectrum of scenarios. It may include structured elements (e.g., numerical vectors) and unstructured elements (e.g., text, images).

\subsection{Representation Learning Process}

A representation learning function \( f \) maps each predictor \( x \in \mathcal{X} \) to a \( p \)-dimensional representation vector \( z \in \mathcal{R}^{p} \), i.e., \( f: \mathcal{X} \rightarrow \mathcal{R}^{p}, \, x \mapsto z \).
Representation learning methods vary based on the type of input data and the analyst’s preferences. For structured data, techniques such as principal component analysis (PCA) are commonly used for dimensionality reduction. For image data, methods like convolutional autoencoders (AE) or contrastive learning approaches (e.g., SimCLR \citep{chen2020simple}) have proven effective in extracting meaningful representations. For text data, models such as Word2Vec or transformer-based approaches (e.g., BERT) are frequently employed to generate semantic embeddings. Moreover, each method involves specific hyperparameters (e.g., the number of principal components in PCA, network architecture details in autoencoder or transformer models) that can lead to variations in the learned representations, thereby constituting different representation learning techniques.

We consider \( M \) representation learning models as our FRL models, each mapping the input space \(\mathcal{X}\) to a model-specific \( p_m \)-dimensional vector space \(\mathcal{R}^{p_m}\), i.e., for the \( m \)-th model,
\[
f_m: \mathcal{X} \rightarrow \mathcal{R}^{p_m}, \quad x \mapsto z_{(m)}.
\]
The function space is defined by $\mathcal{F}$.
These models are trained on the unlabeled dataset \(\mathcal{D}^{\text{u}} = \{x_i^{\prime} \in \mathcal{X}\}_{i=1}^N\), where $x_i^{\prime}$ is the $i$-th unlabeled sample and \( N \) denotes the number of unlabeled samples. Suppose the \( m \)-th FRL model employs a representation learning loss \( U_m(\cdot) \) on \(\mathcal{D}^{\text{u}}\) to obtain
\[
\widehat{f}_m = \textstyle\arg\min_{f\in \mathcal{F}} \sum_{i=1}^N U_m(f(x_i^{\prime})).
\] 
For example, let us illustrate the PCA process to extract representations for panel data.  Define the data matrix by $X^{\prime} = (x_1^{\prime},\ldots,x_N^{\prime})^{\top} \in \mathcal{R}^{N\times T}$. PCA seeks a linear map that gets the $p_m$-dimensional latent factors:  $z_{i,(m)}^{\prime} = f_m(x_i^{\prime})= B_{m}^{\top} x_i^{\prime} $, where  $B_{m} \in \mathcal{R}^{T \times p_m}$ is the parameter matrix. Let $Z_{(m)}^{\prime} = (z_{1,(m)}^{\prime},\ldots,z_{n,(m)}^{\prime})^{\top}$. Then $\widehat{f}_m(x) = \widehat{B}_{(m)}^{\top}x$ for any vector $x\in \mathcal{X}$, where $\widehat{B}_{(m)}$ is calculated by solving a constrained least squares loss function after centering the columns of $X^{\prime}$:
\begin{align*}
    &\qquad (\widehat{Z}^{\prime}_{(m)}, \widehat{B}_{m}) 
     = \textstyle\arg \min_{B,Z}   \big(\|X - Z B^{\top} \|_F^2 \big), \\
    &  \text{s.t. } T^{-1}\textstyle\sum_{i=1}^N z_{i} z_{i}^{\top} = I_{p_m}  \text{ and } B^{\top}B \text{ diagonal},
\end{align*}
where $z_i$ is the $i$-th row of $Z$ and $\widehat{Z}^{\prime}_{(m)}$ is the estimator of $Z_{(m)}^{\prime}$.
The loss $U_m(\cdot)$ can be a $U$-embedding loss function designed for unannotated data  \citep{dai2022embedding} and can be tailored for different scenarios including contrastive loss or generative modeling objectives \citep{chen2020simple,he2020momentum,kingma2013auto}. In each case, the structure of $f_m$ and the dimension $p_m$ in the resulting representations might differ.


\subsection{Downstream Task Analysis}
In this subsection, we focus on the downstream task analysis on a labeled dataset $D^l = \{ (x_i,y_i) \in \mathcal{X} \times \mathcal{Y}\}_{i=1}^n$, where $n$ is  the number of labeled samples. Specifically, this subsection consists of two parts: constructing a set of candidate representations and developing a model averaging predictor to estimate the output for a new input $x_{\mathrm{new}} \in \mathcal{X}$.

With $M$ learned FRL models $\{\widehat{f}_{1},\widehat{f}_{2}, \ldots, \widehat{f}_{M}\}$, each $\widehat{f}_{m}$ maps $x_i \in \mathcal{X}$ to a vector in $\mathcal{R}^{p_m}$. Symbolically, 
\begin{align*}
    \widehat{z}_{i,(m)} = \widehat{f}_{m} (x_i) \ \text{for } i=1,\ldots,n, \ , m=1, \ldots,M,
\end{align*}
where $\widehat{z}_{i,(m)}\in \mathcal{R}^{p_m}$ is defined as the foundation representation vector under the $m$-th FRL model. By collecting these vectors for all $n$ samples, we form the foundation representation matrix  $\widehat{Z}_{(m)} = (\widehat{z}_{1,(m)}, \ldots, \widehat{z}_{n,(m)})^{\top} \in \mathcal{R}^{n \times p_m}$ for the $m$-th FRL model.  

Next, we introduce candidate representation sets that include both the foundation representations directly learned by FRL models and additional representations obtained through fusion or combinatorial techniques. To avoid notational confusion, with a total of $J$ candidate representation sets ($J\ge M$), for $1\le i \le n$, we denote the $j$-th candidate representation vector by $\widehat{z}_{i,[j]}$ and their corresponding matrices $\widehat{Z}_{[j]} = (\widehat{z}_{1,[j]},\ldots,\widehat{z}_{n,[j]})^{\top}$ for $1 \le j \le J$. Note that the subscript bracket for notations of candidate representation sets is $[]$ and is different from $()$ for the foundation representation set $\widehat{z}_{i,(m)}$ for $1\le m \le M$.   Let $p_j$ be the dimension of the vector $\widehat{z}_{i,[j]}$ and therefore $\widehat{Z}_{[j]} \in \mathcal{R}^{n \times p_j} $.

For additional representation sets, one simple way is to consider a simple fusion technique that concatenates the foundation representations: $ (\widehat{z}_{i,(1)}^{\top},\widehat{z}_{i,(2)}^{\top},\ldots,\widehat{z}_{i,(M)}^{\top})^{\top} \in \mathcal{R}^p$, where $p = \sum_{m=1}^M p_m$. The corresponding fusion matrix is $(\widehat{Z}_{(1)},\widehat{Z}_{(2)},\ldots,\widehat{Z}_{(M)})$. This approach is common in deep learning to integrate complementary information from various models \citep{zhang2020JSTSP}. In addition to simple fusion, we provide Example \ref{example: candidate} below to illustrate several methods for constructing additional representation sets.

\begin{example}\label{example: candidate}
\begin{enumerate} 
    \item Foundation Set Combinations. For instance, when $M=3$ and each $p_m=2$, using all non-empty subsets of the three foundation representation sets yields $J=2^M-1=7$ candidate representation sets. This includes individual foundation sets, pairwise concatenations, and the complete concatenation of all three. Mathematically,  the candidate representation sets are $\widehat{Z}_{[j]} = \widehat{Z}_{(j)}$ for $j=1,2,3$; $\widehat{Z}_{[4]}= (\widehat{Z}_{(1)}, \widehat{Z}_{(2)})$; $\widehat{Z}_{[5]}= (\widehat{Z}_{(1)}, \widehat{Z}_{(3)})$; $\widehat{Z}_{[6]}= (\widehat{Z}_{(2)}, \widehat{Z}_{(3)})$; and $\widehat{Z}_{[7]}= (\widehat{Z}_{(1)}, \widehat{Z}_{(2)}, \widehat{Z}_{(3)})$. 
    

    \item Fusion Matrix Columns Combinations. For instance, when $M=3$ and each $p_m=2$, concatenating all foundation representations forms a fusion matrix with  $p=6$ columns. Taking every non-empty subset of these columns gives $J = 2^p - 1 = 63$ candidate representation sets. Each combinatorial subset of the six columns in the fusion matrix forms a candidate representation set. Although this method yields a larger set of candidates, it also poses challenges in terms of computational complexity and potential redundancy.
    
    \item Domain knowledge-based selection. This method emphasizes the integration of expert insights to select or weight candidate representations. For example, in asset pricing, well-established asset factors may be combined with factors learned via deep learning to enhance interpretability and performance \citep{Feng2024JFQA}. This approach helps mitigate the combinatorial explosion in candidate sets by narrowing the focus to the most relevant representations.  
 
\end{enumerate}
\end{example}

After we construct the $J$ candidate representation sets, we build the relationship between candidate representations and the output on the labeled dataset $\mathcal{D}^l$.  For the $j$-th candidate representation set, a function $g_j$ that maps $\mathcal{R}^{p_j}$ onto $\mathcal{Y}$ is:
\begin{align*}
    g_j: \widehat{z}_{i,[j]} \mapsto y_i \text{ and } g_j \in \mathcal{G},
\end{align*}
where $\mathcal{G}$ is the class of considered functions. The model $g_j$ could be a linear regression, a logistic regression, or a more complex nonlinear model depending on the problem types. The function $g_j$ is learned using $(\widehat{z}_{i,[j]}, y_i)_{i=1}^n$ and the corresponding estimated function is defined by $\widehat{g}_j$ for $1\le j \le J$. When the dimensionality of the representations is too large, $g_j$ can be a sparsity-inducing function to reduce the dimensionality. For example, the regularization method employed in \cite {Li2022JASA} can be utilized.

Consider a new observation $x_{\mathrm{new}}\in \mathcal{X}$ and our goal is to predict its label $y_{\mathrm{new}}$. We first generate $M$ foundation representation sets by applying $M$ FRL models on $x_{\mathrm{new}}$, and then construct  $J$ candidate representation sets for $x_{\mathrm{new}}$, denoted as $\{\widehat{z}_{\mathrm{new},[j]}\}_{j=1}^J$. For each $1\le j \le J$, the candidate model generates an output as follows:
\begin{align*}
    \widehat{y}_{\mathrm{new},[j]} =  \widehat{g}_j(  \widehat{z}_{\mathrm{new},[j]}).
\end{align*}
We propose a model averaging approach to aggregate these $J$ outputs by assigning weights to each of them. Specifically, the model averaging predictor is defined as: 
\begin{align} \label{eq:MAy}
\widehat{y}_{\mathrm{new}}(\bw) = \textstyle\sum_{j = 1}^{J} w_j \widehat{y}_{\mathrm{new},[j]},
\end{align}
where $w_j$ is the weight assigned to the $j$-th output, and the weight vector $\bw = (w_1,\ldots,w_{J})^{\rm T}$ lies in the simplex $\bw$ satisfies that   $\bw  \in \mathcal{W} = \{\bw \in [0,1]^{J}: \sum_{j=1}^{J}w_j =1 \}$. The weight assignment for diverse candidate models is critical in model averaging, as it reflects the relative predictive power of each candidate model.  Next, we discuss how to tune the weights for different downstream tasks.

\subsection{Weight Choice Criterion}
For downstream tasks, the criteria for model evaluation and their corresponding loss functions often differ. 
Our paper aims to build a generalized framework that can adaptively handle different kinds of loss functions based on various machine-learning-based candidate models. A simple but naive approach is to construct the weight criterion that is directly based on the targeted loss function. 
The weights are tuned on the labeled dataset by
\begin{align}\label{eq:wdirect}
    \widehat{\bw} = \text{argmin}_{\mathbf{w} \in \mathcal{W}} n^{-1} \textstyle\sum_{i=1}^n L(y_i, \widehat{y}_i(\bw)),
\end{align}
where $\widehat{y}_i(\bw) = \sum_{j=1}^J w_j\widehat{y}_{i,[j]}$ and $L(\cdot,\cdot)$ is a loss function that measures the difference between the true responses and the weighted predictions.
However, the weighting criterion \eqref{eq:wdirect} has two problems. First, many loss functions, particularly in complex or classification problems, can be non-convex or non-differentiable (e.g., the 0-1 loss or certain robust loss functions). These properties hinder the direct application of gradient-based optimization methods, potentially leading to suboptimal solutions or convergence issues.  
Second, optimizing the loss over the all training samples may cause the model to fit the noise or idiosyncratic patterns in the training data rather than capturing the underlying general patterns. This may increase the risk of overfitting problem \citep{john2010elements}.

For the computational complexity problem in \eqref{eq:wdirect}, surrogate loss functions that are convex and differentiable are often employed.  For some surrogate losses, their connection to the original loss is supported by theoretical guarantees, such as classification calibration, consistency in risk minimization, or upper bounds on the excess risk measured by the original loss \citep{mao2023ICML}. For example, in classification task, the hinge loss \citep{cortes1995support}, the logistic loss  \citep{zhu2005kernel}, and the $\psi$-loss \citep{shen2003psi}  are often employed as smooth approximations for the zero-one classification loss. These surrogate losses not only facilitate optimization but also have been shown to preserve desirable theoretical properties under certain conditions, such as minimizing the surrogate loss leads to minimization of the original loss under suitable conditions.
In addition, if the original loss function is computational friendly, such as the mean squared error in regression tasks, it is entirely reasonable to use the original loss function without substitution.

For the overfitting problem in \eqref{eq:wdirect}, we propose a cross-validation-based weight tuning algorithm (Algorithm \ref{algo:cvma}). Let $S=\{1,\ldots,n\}$ denote the index set, which is divided into $K$ equally sized folds: $S=\bigcup_{k=1}^K S_k$, where $S_k$ is the index set in the $k$-th fold. For the $j$-th candidate model and each $k\in\{1,\ldots,K\}$, we refit the model $g_j$ using the dataset $\{(\widehat{z}_{i,[j]}, y_i) \}_{i\in S \setminus S_k}$. The resulting refitted function of $g_j$ excluding the $k$-fold is denoted by $\tilde{g}^{(-k)}_{j}$. 
Then $\tilde{g}^{(-k)}_{j}$ is applied on $\{\widehat{z}_{i,[j]}\}_{i\in S_k}$ to obtain $\tilde{y}_{i,[j]} = \tilde{g}^{(-k)}_{j}(z_{i,[j]})$ with $i \in S_k$. By aggregating the predictions from all folds, we get the cross-validation-based prediction for all labeled samples under the $j$-th candidate model, denoted by $\widetilde{y}_{[j]} = (\widetilde{y}_{1,[j]}, \ldots, \widetilde{y}_{n,[j]})^{\top}$. Finally, we estimate the weights by 
\begin{align}\label{eq:weights}
    \widehat{\bw} = \text{argmin}_{\mathbf{w} \in \mathcal{W}} n^{-1} \textstyle\sum_{i=1}^n V (y_i, \widetilde{y}_{i}(\bw)),
\end{align}
where $V$ is the surrogate loss function for the original loss function $L$, $\widehat{\bw} = (\widehat{w}_1,\ldots,\widehat{w}_J)^{\top}$, and $\widetilde{y}_{i}(\bw) = \sum_{j=1}^J w_j \widetilde{y}_{i,[j]}$.

By \eqref{eq:weights}, we plug in the obtained $\widehat{\bw}$ in \eqref{eq:MAy} to derive our PEARL prediction for the label of the new observation $x_{\mathrm{new}}$:
\begin{align}\label{predict:MA}
    \widehat{y}_{\mathrm{new}}(\widehat{\bw}) = \textstyle\sum_{j = 1}^{J} \widehat{w}_j \widehat{y}_{\mathrm{new},[j]}.
\end{align}

\begin{algorithm}
\caption{$K$-Fold cross-validation weight criterion}
\label{algo:cvma}
\begin{algorithmic}[1] 
    \State \textbf{Input:} Labeled data $D^l = \{(x_i,y_i)\}_{i=1}^n$; Number of folds $K$; $M$ FRL models $\{\widehat{f}_1,\ldots,\widehat{f}_M \}$;  Surrogate loss function $V$.  
    \State Construction of the representations: For $m\in\{1,...,M\}$ and $i\in\{1,...,n\}$, $\widehat{z}_{i,(m)} = \widehat{f}_m(x_i)$; for $j\in\{1,...,J\}$ generate $\widehat{z}_{i,[j]}$.
    \State Cross-validation weight tuning: Partition index set $S=\{1,\dots,n\}$ into $K$ folds: $S=\bigcup_{k=1}^K S_k$.
    \For{$j \leftarrow 1$ \textbf{to} $J$}; \For{$k \leftarrow 1$ \textbf{to} $K$}
          \State $D_{\text{train}} \gets \{(\widehat{z}_{i,[j]}, y_i) \}_{i\in S \setminus S_k}$;
          \State Train the prediction model $\tilde{g}^{(-k)}_{j}$ on $D_{\text{train}}$;
          \State Obtain $\widetilde{y}_{i,[j]} = \tilde{g}^{(-k)}_{j}(\widehat{z}_{i,[j]})$ for $i \in S_k$.
        \EndFor
        \State Summarize all folds: $\widetilde{y}_{[j]}= (\widetilde{y}_{1,[j]},\ldots,\widetilde{y}_{n,[j]} )^{\top}$.
    \EndFor
    \State The weighted prediction is given by
    $\widetilde{y}_{i}(\bw) = \sum_{j=1}^J w_j \widetilde{y}_{i,[j]}$.
    where $\bw = (w_1,\ldots,w_{J})^{\rm T}\in\mathcal{W}$ 
    \State Solve for the weights by minimizing
    \begin{align*}
        \widehat{\bw} = \text{argmin}_{\mathbf{w} \in \mathcal{W}} \frac{1}{n} \sum_{i=1}^n V (y_i, \widetilde{y}_{i}(\bw)).
    \end{align*}
    \State \textbf{Output:} $\widehat{\bw}$.
\end{algorithmic}
\end{algorithm}

\subsection{PEARL for Multimodal Data}

This part illustrates how the proposed method PEARL is applied to multimodal representation learning. In many practical applications, data are available in multiple modalities—such as images and text for news articles or sequences of images, audio, and text for YouTube videos. The proposed method PEARL can effectively combine representations learned from each modality.

Suppose our predictor consists of \( S \) distinct modalities, abstractly represented as
\[
\mathcal{X} = \mathcal{X}^1 \cup \mathcal{X}^2 \cup \cdots \cup \mathcal{X}^S,
\]
where the union symbol is used in an abstract sense. In practice, integrating these heterogeneous data sources may require a Cartesian product or another specific form of fusion, depending on how the features interact. We consider a large collection of \( N \) unlabeled samples, $\mathcal{D}^u = \{ x_i' \in \mathcal{X} : i = 1, \ldots, N \}$, to train a total of \( M \) foundational representation learning (FRL) models. These FRL models are designed to learn representations from the multimodal input data, and they can operate on single modalities or combinations thereof. For illustration, consider two example FRL models for the input $x=(x^1,\ldots,x^S)\in\mathcal{X}$:
\[
\begin{aligned}
    & f_1: \mathcal{X}^1 \rightarrow \mathcal{R}^{p_1}, \quad x^1 \mapsto z_{(1)}, \\
    & f_2: (\mathcal{X}^2, \mathcal{X}^3) \rightarrow \mathcal{R}^{p_2}, \quad (x^2,x^3) \mapsto z_{(2)}.
\end{aligned}
\]
Here, \(f_1\) represents FRL models that focus on individual modalities, while \(f_2\) represents those built to capture joint information from a combination of modalities (see, e.g., \citep{Ngiam2011ICML, Shekhar2014IEEE, gu2017learning}). The complete set of trained FRL models is denoted as \(\{\widehat{f}_1, \ldots, \widehat{f}_M\}\).

For the downstream task, we assume that there are \( n \) labeled samples forming the dataset $\mathcal{D}^l = \{ (x_i, y_i) \in \mathcal{X} \times \mathcal{Y} : i = 1, \ldots, n \}$
where each \( x_i = (x_i^1, x_i^2, \ldots, x_i^S) \) is a multimodal input and \( y_i \) is the corresponding label. For example, the foundational representations for sample \( x_i \) obtained via the first two FRL models are $\widehat{z}_{i,(1)} = \widehat{f}_1(x_i^1)$ and $\widehat{z}_{i,(2)} = \widehat{f}_2(x_i^2, x_i^3)$, respectively.
After extracting representations from all \( M \) FRL models, we generate \( J \) candidate representation sets, denoted by \(\widehat{z}_{i,[j]}\) for \( i = 1, \ldots, n \) and \( j = 1, \ldots, J \). The subsequent steps of our methodology follow the procedure outlined in our main algorithm, where we derive cross-validation weights, fit the prediction models, and make prediction for any new observation $x_{\operatorname{new}}\in\mathcal{X}$.

\section{Theoretical Results}\label{TheoreticalResults}
In this section, we demonstrate the theoretical efficiency of our proposed method, PEARL, by analyzing two key aspects. First, we show that the risk of the PEARL predictor in downstream tasks is asymptotically equivalent to the theoretical minimum risk. Second, we establish that the sum of the weights assigned to the correctly specified models converges to one. 

For the representation learning process, let  $f^*_{m}$  represent the optimal representation learning function that maps from $\mathcal{X}$ to $\mathcal{R}^{p_m}$ under the given representation learning loss $U_m(\cdot)$, that is $f^*_{m} = \arg \min_{f \in \mathcal{F}} \mathbb{E}[U_m(f(x))]$ with $x \in \mathcal{X}$ for $1\le m \le M$. 
For $x \in \mathcal{X}$, we then define the optimal foundation representation for $x$ under the $m$-th FRL model as $z_{(m)}^* = f^*_{m}(x)$ for $1 \le m \le M$. By concatenating the optimal representations from all $M$ FRL models, we obtain the fused optimal representations $z^*=(z_{(1)}^{*\top},z_{(2)}^{*\top},\ldots,z_{(M)}^{*\top})^{\top}$ by directly augmenting the optimal representations from FRL models. By following procedure in Example \ref{example: candidate}, we denote the optimal candidate representations by   $z_{[j]}^*$ for $1\le j \le J$ by applying the candidate construction procedure to optimal foundation representations. 

For the downstream task analysis process, for $1 \le j \le J$, 
let $g^*_{j} = \arg\min_{g\in \mathcal{G}} \mathbb{E}[L( y_i, g(z^*_{[j]})) ]$  be the optimal learning function using the $j$-th candidate representation set. Let a function $h^* (x,\bw) = \sum_{j=1}^J w_j g^*_j(z^*_{[j]}) $ be the model averaging value of outputs derived from optimal candidate representations and optimal downstream task learning function, as $z^*_{[j]}$ depends on $x$. Define $\bw^* = (w_1,\ldots,w_J)^{\top}$ that satisfies $\bw^* = \arg \min_{\mathbf{w}\in \mathcal{W}} E[L(y,h^* (x,\bw) ]  $.  Analogously, for the estimated values, define $\widehat{h} (x,\bw) = \sum_{j=1}^J w_j \widehat{g}_j(\widehat{z}_{[j]}) $.

\begin{assumption}\label{assumption:bias_LV} 
Suppose there exist constants $\mu>0$ and $c_1>0$ such that when
 $  \mathbb{E} [V(y, \widehat{h} (x,\bw))]  - \mathbb{E}[ V(y, h^* (x,\bw^*))  ]\le \epsilon$ for $\bw \in \mathcal{W}$ and a  small error $\epsilon>0$, then 
    \begin{align*}
          \mathbb{E}  [ L(y, \widehat{h} (x,\bw))  ] - \mathbb{E} \left[ L(y, h^* (x,\bw^*)) \right] \le c_1 \epsilon^{\mu}.
    \end{align*}  
\end{assumption}
Assumption \ref{assumption:bias_LV} requires that a small bias bound in the surrogate function $V$ can translate into a relatively small bias bound in the true loss function $L$.
The polynomial relationship between the bias bound of $V$ and $L$ is also adapted in  Assumption A1 in \cite{dai2022embedding}. Such conditions hold in many standard machine-learning problems where $V$ is carefully designed to approximate $L$, such that by controlling the bias of the surrogate loss, we also control the bias of the true loss.

\begin{assumption}\label{assumption:var_LV}
   For each $1\le i \le n$ and $\bw \in \mathcal{W}$, if $\mathbb{E} [V(y_i,\widetilde{y}_i(\bw)) - V(y_i, {y}_i^*(\bw))] \le \epsilon$ for some small error $\epsilon>0$, then there exist constants $\kappa>0$ and $c_2>0$ such that,
    \begin{align*}
       \var [V(y_i,\widetilde{y}_i(\bw)) - V(y_i, {y}_i^*(\bw))] \leq c_2\epsilon^{\kappa}.
    \end{align*} 
\end{assumption}
Assumption \ref{assumption:var_LV} bounds the variance of loss difference between $\widetilde{y}_i(\bw)$ and ${y}_i^*(\bw))$ if the expected difference is small in the surrogate function $V$.  This assumption requires a mild smoothness for $V$, where the constant $\kappa$ reflects the smoothness. Assumption \ref{assumption:var_LV} is similar to Assumption A2 in \cite{dai2022embedding}.

\begin{assumption}\label{assumption:varystar}
    For any $\bw \in \mathcal{W}$, $\var (V(y, h^*(x,\bw)) \le c_3 $, where $c_3$ is a positive constant.
\end{assumption}
Assumption \ref{assumption:varystar} requires that variances of the surrogate loss of model averaging estimator derived from optimal representations and downstream task learning function  is bounded by a positive constant. It indicates that the data distribution and the model do not produce arbitrarily large fluctuations in the surrogate loss.

\begin{assumption}\label{assumption:equicontinous}
For two predicted outputs $y{'}$ and $y{''}$, it has $ \vert V(y,y{'}) - V(y,y{''})\vert \le K(y) \| y{'} - y{''}\| $, where $K(y)$ is independent of $y{'}$ and $y{''}$, and satisfies $   \mathbb{E}^{1/2} (\vert K(y) \vert^2) \le c_4$ for some positive constant $c_4$.
\end{assumption}
Assumption \ref{assumption:equicontinous} requires the smoothness of the surrogate function. It is similar to Assumption 2 in \citep{yu2024unif}, which poses the stochastic equicontinuity on the measure of divergence in the original loss function,  and Assumption A in \citep{zhu2016jasa} and Assumption 4.2 in \citep{bi2018aos}, the two of which pose the smoothness on the likelihood function derived the original loss function. Compared with these smoothness assumptions imposed on $L$, Here, we apply it to $V$, thereby broadening the class of possible original losses that lack such smoothness but still allow a suitably chosen
$V$ to meet this requirement. 
This assumption is often key for establishing 
uniform convergence of empirical loss processes, 
as it ensures small differences in the predicted output lead to small changes in the surrogate loss measure.

Assumption \ref{assumption:equicontinous} can be satisfied when the surrogate loss is a \emph{margin-based} loss. For hinge loss $V(y,f) = \max\{ 0,\;1 - y y'\}$ under a binary classification setting with $y \in \{-1,+1\}$, it satisfies that  
\[
   \bigl|\max\{0,\;1-y y'\}
         - \max\{0,\;1-y y''\}\bigr|
   \;\le\;
   \bigl|(1-y y') - (1-y y'')\bigr|
   \;=\;
   \vert y\vert \bigl\vert y' - y''\bigr\vert
   \;\le\;
   \bigl|y' - y''\bigr|,
\]
where $K(y)=1$. For the logistic loss (cross-entropy loss)
$V\bigl(y, y'\bigr) = \log\bigl(1 + e^{- y y'}\bigr)$ with $ y\in\{-1,+1\}$, its derivative with respect to $y'$ satisfies
\[
   \frac{\partial}{\partial y'} 
   \log\!\bigl(1 + e^{- y y'}\bigr)
   \;=\;
   - \frac{y e^{- y y'}}{1 + e^{- y y'}},
\]
whose absolute value is bounded by $1$. Hence, the logistic loss is Lipschitz
in $y'$ with a constant at most $\vert y\vert$ and $K(y)=1$ in Assumption \ref{assumption:equicontinous}. More generally, many margin-based losses take the form 
$V\bigl(y, y'\bigr)=\phi\!\bigl(y y'\bigr)$ for some convex and often smooth function
$\phi$, including the exponential loss, the smoothed hinge loss, and various robust surrogates. In each case, one typically finds a bounded derivative 
with respect to $y'$, thus admitting a finite Lipschitz constant $K(y)$ that 
depends at most on~$\vert y\vert$. For binary classification 
with $y\in\{-1,+1\}$, we immediately get $|y|=1$, yielding a constant $K(y)\le 1$.
In multi-class or other extended settings, one can assume $\|y\|\leq C_y$ 
with high probability, ensuring $\mathbb{E}[K(y)^2] < c_4$.

\begin{assumption}\label{assumption:bias_predict}
    There exists a sequence $\alpha_{N,n}$ depending on $N$ and $n$, such that
 \begin{align*}
     \max_{1 \le j \le J} \mathbb{E}^{1/2} (\|\widehat{g}_j(\widehat{z}_{[j]}) - g_j^* (z^*_{[j]}) \|^2) = O(\alpha_{N,n})
 \end{align*} 
 and 
 \begin{align*}
    \max_{1 \le j \le J} \mathbb{E}^{1/2} (\|\widetilde{g}_j(\widehat{z}_{[j]}) - g_j^* (z^*_{[j]}) \|^2) = O(\alpha_{N,n}).
\end{align*}  
 \end{assumption}

 Assumption \ref{assumption:bias_predict} shows the maximum error bound of $J$ candidate estimators and their corresponding CV estimators to the limiting values for $J$ candidate models. In practice, $\alpha_{N,n}$ may shrink to $0$ as $N,n\to\infty$, 
reflecting the convergence rate of estimators and their corresponding CV estimators. The requirement holds \emph{uniformly} over $j=1,\dots,J$, which is natural in  model-averaging contexts where multiple candidate models are considered.

 The error bound of $\widehat{g}_j(\widehat{z}_{i,[j]})$ can be decomposed by the approximation error (the discrepancy introduced by training downstream learning function) and the estimation error (the discrepancy introduced by estimating representations) as follows
 \begin{align}\label{eq:decomp}
   \quad \|\widehat{g}_j(\widehat{z}_{i,[j]}) - g^*_j (z^*_{i,[j]}) \|  
      &\le \underbrace{\| \widehat{g}_j({z}^*_{i,[j]}) - g^*_j (z^*_{i,[j]}) \|}_{\text{approximation error}} + \underbrace{\|\widehat{g}_j(\widehat{z}_{i,[j]}) - \widehat{g}_j({z}^*_{i,[j]}) \|}_{\text{estimation error}}.
 \end{align}
The approximation error depicts the ability of model $\widehat{g}_j$ in Step 2 to approximate the optimal model  $g^*_j$.  The approximation error can potentially be reduced by using more flexible model specifications, which allow experts to better tailor the model to the data. However, increasing model flexibility also introduces the risk of overfitting. Overfitting occurs when the model becomes too closely aligned with the specific patterns in the training data, which may lead to poor performance when applied to new and unseen data.
The estimation error arises due to the representation learning process in Step 1. If the used function $\widehat{g}$ falls in the Lipschitz-continuous function space, then the estimation error is bounded by $c^L \|\widehat{z}_{i,[j]} -{z}^*_{i,[j]} \|$ with a positive Lipschitz constant $c^L$.
Analogously, the estimation error of $\widetilde{g}_j(\widehat{z}_{i,[j]})$ can be decomposed similar to \eqref{eq:decomp} and it is assumed that the cross validated estimator inherits the convergence rate of the original estimator.

In addition, the framework can be extended to encompass commonly used pretrained models as foundation models. Unlike the assumption in the original formulation, where all foundation models are trained on the same unlabeled dataset, these pretrained models may be trained on different unsupervised datasets. Subsequently, they use the learned representations for downstream task analysis. Specifically, let $\mathcal{D}^u_m$ denote the unsupervised dataset for the $m$-th foundation model, with a sample size of $N_m$. To generalize the framework for these pretrained models, we modify Assumption \ref{assumption:bias_predict} by setting $N = \max_{1 \le m \le M} N_m$.

\subsection{Asymptotic Optimality}

Define the risk function of model averaging predictor with respect to the label of a new observation $x_{\mathrm{new}}$ by
\begin{align} \label{eq:RL}
        R_L(\bw) 
        = \mathbb{E}\big\{L(y_{\mathrm{new}},\widehat{y}_{\mathrm{new}}(\bw))\big\} 
        =\mathbb{E} \big\{L(y_{\mathrm{new}}, \sum_{j = 1}^{J} {w}_j \widehat{g}_j(\widehat{z}_{\mathrm{new},[j]}))\big\}.
\end{align}
Now, let's define the risk function of the weighted predictor using the limiting values of each model's prediction:
\begin{align} \label{eq:RL1}
    R_L^*(\bw) = \mathbb{E}\big\{L(y_{\mathrm{new}}, {y}^*_{\mathrm{new}}(\bw))\big\} 
    =\mathbb{E} \big\{L(y_{\mathrm{new}}, \sum_{j = 1}^{J} {\omega}_j  {g}^*_j( {z}^*_{\mathrm{new},[j]}))\big\}.
\end{align}
We define the surrogate risk function $R_V(\bw)$ of our model averaging estimator by replacing $L(\cdot)$ with $V(\cdot)$ in \eqref{eq:RL}. The corresponding risk function with limiting values $R_V^*(\bw)$ is then defined by replacing $L(\cdot)$ with $V(\cdot)$ in \eqref{eq:RL1}. Define the infimum risk of $R_L^*(\bw)$ by $\xi_L = \inf_{\mathbf{w} \in \mathcal{W}} R_L^*(\bw)$, and the infimum risk of $R_V^*(\bw)$ by $\xi_V = \inf_{\mathbf{w} \in \mathcal{W}} R_V^*(\bw)$.

\begin{assumption}\label{assumption:opt_main}
    $\xi_V^{-1} \xi_L =O(1)$ and $\xi_V^{-1} \max \{\alpha_{N,n}^{\kappa/2} n^{-1/2}, \alpha_{N,n}^{\kappa}, \alpha_{N,n}^{\mu} \}=o(1)$.
\end{assumption}
Assumption \ref{assumption:opt_main} contains two parts. Part $I=\xi_V^{-1} \xi_L =O(1)$ keeps minimal possible surrogate risk on the same order with the minimal possible true loss. It is a mild or non-degenerate condition ensuring that improvements on the surrogate side can be translated to improvements on the true loss. To put it in another words, if $\xi_V$ is extremely larger than $\xi_L$, it may imply that even the best surrogate solution performs poor in the true loss, making it pointless to minimize $V$ instead of $L$.
Part $II =\xi_V^{-1} \max \{\alpha_{N,n}^{\kappa/2} n^{-1/2}, \alpha_{N,n}^{\kappa}, \alpha_{N,n}^{\mu} \}=o(1) $ means that the minimal possible surrogate risk $\xi_V$ diverges at the rate faster than $\max \{\alpha_{N,n}^{\kappa/2} n^{-1/2}, \alpha_{N,n}^{\kappa}, \alpha_{N,n}^{\mu} \}$. The term $\alpha_{N,n}$ measures the approximation error and estimation error of the two stages, and the term $\alpha_{N,n}^{\kappa/2} n^{-1/2}$ arises by controlling the cross-validation and sample-average difference under polynomial variance bounds, and $\alpha_{N,n}^{\mu}$ arises by converting the risk difference of surrogate loss to that of the true loss.

\begin{theorem}  \label{theorem:opt}
    Under Assumptions \ref{assumption:bias_LV}-\ref{assumption:opt_main}, we have
    \begin{align*}
    \frac{R_L(\widehat{\bw})}{\inf_{\mathbf{w}\in \mathcal{W}} R_L(\bw)} \xrightarrow{p} 1.
    \end{align*}
\end{theorem}

Theorem \ref{theorem:opt} shows the asymptotic optimality of PEARL predictor of $y_{\mathrm{new}}$ in terms of risk. It means that PEARL predictor can asymptotically outperform or at least match any convex combination of candidate models when $\bw \in \mathcal{W}$. A commonly used convex combination is the simple model averaging with equal weights, $\bw = (M^{-1}, \ldots, M^{-1})^{\top}$. Additionally, the PEARL predictor can also outperform or match any single candidate model, as the weight vector with one element equal to one and others zero lies in  $\mathcal{W}$. The details of the theoretical assumptions and proof can be found in the Appendix.

\subsection{Weight Consistency}
Next, we prove the consistency of the weights that the estimated weights are assigned to the correctly specified models when such models are included in the averaging pool. The concept of the  correctly specified candidate models refers that $g^*_{j}(z^*_{\mathrm{new},[j]}) = \mathbb{E}(y_{\mathrm{new}}|x_{\mathrm{new}})$ for the $j$-th candidate model. 

Let $\mathcal{D} \subset \{1,\ldots,J\}$ be a  subset of $\{1,\ldots,J\}$ that contains all the index of the  correctly specified candidate models. Define $\widehat{\tau} = \sum_{j \in \mathcal{D}} \widehat{w}_j$ be the sum of weights assigned to the  correctly specified candidate models. Assume that that $\mathcal{D}$ is not empty, and it means that at least one correctly specified model is included in the candidate model set, and therefore we only need to prove that $\widehat{\tau} \rightarrow 1$ to show the consistency of weights. In addition, since if all candidate models are correctly specified, then the result $\widehat{\tau} =1$ is directly derived, we assume that $\mathcal{D} \neq \{1,\ldots,J\}$. 
Define $\mathcal{W}_s = \{\bw \in \mathcal{W}: \sum_{j \notin \mathcal{D}} w_j = 1 \}$ be a weight set that assigns all weights on the misspecified models .

\begin{assumption}\label{assumption:RVL2}
    For every $\bw \in \mathcal{W}$, there are  
      $R_V^*(\bw) \ge   c_5 \|  y^*_{\mathrm{new}} (\bw)   - \mathbb{E}(y_{\mathrm{new}}|x_{\mathrm{new}})\|^2 $,
    for some positive constant $c_5$.
\end{assumption}
Assumption \ref{assumption:RVL2} is similar to Assumption 5 in \citep{yu2024unif} which assumes the lower bound on the risk of true loss, and we focus on the surrogate one. Next, let us show how commonly used surrogate losses—squared error, logistic (cross-entropy), and hinge loss  imply Assumption \ref{assumption:RVL2}.
For squared-error loss where $V(y_{\mathrm{new}}, y^*_{\mathrm{new}}(\bw)) = L(y_{\mathrm{new}}, y^*_{\mathrm{new}}(\bw)) = (y^*_{\mathrm{new}}(\bw) - y_{\mathrm{new}} )^2$,
By the usual orthogonality argument for squared error, 
$$ 
  R_V^*(\bw) 
  = \mathbb{E}\bigl[
   (y^*_{\mathrm{new}}(\bw) - y_{\mathrm{new}} )^2 \bigr]
  = \mathbb{E}\bigl[ (y^*_{\mathrm{new}}(\bw)
      - \mathbb{E}[y_{\mathrm{new}}\mid x_{\mathrm{new}}] )^2 \bigr]
  +
  \mathbb{E}\bigl[ (\mathbb{E}[y_{\mathrm{new}}\mid x_{\mathrm{new}}] - y^*_{\mathrm{new}}(\bw))^2 \bigr].
$$
Therefore, we have  
$R_V^*(\bw) \ge
  \bigl\| y_{\mathrm{new}}^*(\bw) - \mathbb{E}[ y_{\mathrm{new}}\mid x_{\mathrm{new}}] \bigr\|^2$.
  
For logistic (cross-entropy) loss with $y_{\mathrm{new}}\in\{-1,1\}$, we know  
$ V\bigl(y_{\mathrm{new}}, y^*_{\mathrm{new}}(\bw)\bigr) = \log\bigl(1+ \exp(- y_{\mathrm{new}} y^*_{\mathrm{new}}(\bw))\bigr)$ and  $R_V^*(\bw) = \mathbb{E}[
    \log(1 + \exp\{-y_{\mathrm{new}}y^*_{\mathrm{new}}(\bw)\})]$. 
 Since $ \ell(m) = \log (1 + e^{m})$
is strictly convex relative to $m$. Then
\begin{align*}
  & \quad \log\bigl(1+ \exp(- y_{\mathrm{new}} y^*_{\mathrm{new}}(\bw))\bigr)
  -
   \log\bigl(1+ \exp(- y_{\mathrm{new}} \mathbb{E}(y_{\mathrm{new}} |x_{\mathrm{new}} )\bigr) \\
  &\ge
  c\bigl(
    y_{\mathrm{new}} \mathbb{E}(y_{\mathrm{new}} |x_{\mathrm{new}} ) - y_{\mathrm{new}} y^*_{\mathrm{new}}(\bw)
  \bigr)^2 \\
  &\ge c\bigl(
   \mathbb{E}(y_{\mathrm{new}} |x_{\mathrm{new}} ) -  y^*_{\mathrm{new}}(\bw) \bigr)^2
\end{align*}
for some constant $c>0$. Then we have
\begin{align*}
  R_V^*(\bw)
  - \mathbb{E}\Bigl[ \log\bigl(1 + e^{- y_{\mathrm{new}} \mathbb{E}(y_{\mathrm{new}} |x_{\mathrm{new}} )}\bigr) \Bigr]
 \ge
  c \mathbb{E}\Bigl[
    \bigl\vert y^*_{\mathrm{new}}(\bw) 
      -\mathbb{E}(y_{\mathrm{new}} |x_{\mathrm{new}} ) \bigr\vert^2
  \Bigr.
\end{align*}
Since the term $[ \log\bigl(1 + e^{- y_{\mathrm{new}} \mathbb{E}(y_{\mathrm{new}} |x_{\mathrm{new}} )}\bigr) ]$ is always larger than 0, then the logistic loss satisfies Assumption \ref{assumption:RVL2}.

For hinge loss with $y_{\mathrm{new}}\in\{-1,+1\}$, we have
$  V\bigl(y_{\mathrm{new}}, y_{\mathrm{new}}^*(\bw)\bigr)
=
  \max\{0,1 - y_{\mathrm{new}}y_{\mathrm{new}}^*(\bw)\}$.
By piecewise linearity and local strong-convexity that
$$
   R_V^*(\bw)
 -
  \mathbb{E}[V(y_{\mathrm{new}}, \mathbb{E}(y_{\mathrm{new}}|x_{\mathrm{new}})]
 \ge
  c \mathbb{E}[\bigl\| y_{\mathrm{new}}^*(\bw) - \mathbb{E}(y_{\mathrm{new}}|x_{\mathrm{new}})\bigr\|^2],
$$
where the term $\mathbb{E}[V(y_{\mathrm{new}}, \mathbb{E}(y_{\mathrm{new}}|x_{\mathrm{new}})]$ is a constant that does not depend on $\bw$. Therefore, the hinge loss still satisfies Assumption \ref{assumption:RVL2}.

\begin{assumption}\label{assumption:weight_xi}
    $\max\{\alpha_{N,n}^{\kappa/2}n^{-1/2}, \alpha_{N,n}^{\kappa}\} \{\inf_{\mathbf{w} \in \mathcal{W}_s } \| \sum_{j \notin \mathcal{D} }  {w}_j [y_{i,[j]}^*  - \mathbb{E}(y_{\mathrm{new}}|x_{\mathrm{new}})]  \|^2\}^{-1} = o(1)$.
\end{assumption}
Assumption \ref{assumption:weight_xi} means that the divergence speed of minnimum risk of averaging across all misspecified models is faster than $\max\{\alpha_{N,n}^{\kappa/2}n^{-1/2}, \alpha_{N,n}^{\kappa}\}$. 

\begin{theorem}\label{theorem:weight}
    If Assumptions \ref{assumption:bias_LV}-\ref{assumption:bias_predict}  and \ref{assumption:RVL2}-\ref{assumption:weight_xi} are satisfied, then we have $\widehat{\tau} \rightarrow 1$.
\end{theorem}
Theorem \ref{theorem:weight} shows that the nonzero weights will asymptotically be put on the correctly specified candidate models when at least one correctly specified model exists in the candidate set. This result implies that, PEARL will asymptotically concentrate on the models whose optimal predicted values coincide with the conditional expectation.

\section{Experiments} \label{Experiments}
This section presents a simulation study and four real-world datasets to demonstrate the effectiveness of PEARL across different tasks and domains. The simulation study tests a non-linear regression data generating process for the regression task. Four real-world datasets consist of MNIST \citep{deng2012mnist} and CIFAR-10~\citep{krizhevsky2009learning} for image classification, Polarity \citep{PangLee04a} for sentiment analysis, and Wine Review\footnote{\url{https://www.kaggle.com/datasets/zynicide/wine-reviews}} for multi-modal learning. A summary of experiments and their corresponding FRL methods is presented in Table~\ref{tab:methods_summary}, with detailed implementation information provided in Appendix B.  For classification tasks in our experiments, we use the cross entropy (CE) loss as the surrogate function $V$, and other tasks use the squared loss function when calculating the weights. 

\begin{table}[ht]
\centering
\caption{Summary of FRL methods applied to different tasks and datasets.}
\label{tab:methods_summary}
\begin{tabular}{>{\raggedright}p{8cm} l}
\toprule
\textbf{Task Type \& Dataset}  & \textbf{FRL Methods}             \\
\midrule
\multirow{5}{*}{\parbox[t]{8cm}{Synthetic dataset: A synthetic regression task with a nonlinear data generating process.}} & PCA \\
                     & KPCA (Gaussian Kernel) \\
                     & KPCA (Polynomial Kernel) \\
                     & KPCA (Sigmoid Kernel) \\
                     & KPCA (Cosine Kernel) \\
\midrule
\multirow{5}{*}{\parbox[t]{8cm}{MNIST: An image classification task that classifies handwritten digits into one of ten categories.}} & AE (CNN encoder) \\
                   & VAE (CNN encoder) \\
                     & SimCLR (MLP encoder)  \\
                     & SimCLR (CNN encoder) \\
                     & Original Image \\
\midrule
\multirow{4}{*}{\parbox[t]{8cm}{CIFAR-10: An image classification task that classifies color images into one of ten categories.}} 
& AE (ResNet encoder) \\
                   & VAE (ResNet encoder) \\
                     & SimCLR (CNN encoder)  \\
                     & SimCLR (ResNet encoder) \\
\midrule 
\multirow{6}{*}{\parbox[t]{8cm}{Polarity: A text classification task that predicts sentiment polarity (positive or negative) from movie reviews.}} & CountVectorizer \\
                    & TF-IDF Vectorizer \\
                     & Word2Vec \\
                     & Doc2Vec  \\
                     & GloVe \\
                     & BERT  \\
\midrule
\multirow{4}{*}{\parbox[t]{8cm}{Wine Review: A multi-modal regression task that predicts the quality of wines.}} & PCA (Numerical Modality)\\
                     & VAE (Numerical Modality) \\
                     & BERT (Textual Modality) \\
                     & GPT-2 (Textual Modality) \\
\bottomrule
\end{tabular}
\end{table}

We compare PEARL with following baseline methods:  {Best}, {Fusion}, {SA-FRL}, {SA-cand}, and {MS}. Each of these methods provides a different strategy for aggregating or selecting models. {Best} reports the best individual FRL model with the best performance, serving as a benchmark to assess whether leveraging multiple FRL models improves performance. {Fusion} refers to the direct fusion method that uses the combination of representations of all considered FRL models. Fusion shows the result of using all representations without weighting any individual models. {SA-FRL} is a simple averaging method that equally averages predictions of all individual FRL models. {SA-cand} is another simple averaging method that equally averages the predictions of all candidate models. The difference between SA-FRL and SA-cand lies in the set of models being averaged. {MS} is a model selection method that selects the best candidate model based on cross-validation. MS is compared with PEARL to show the advantages of using a weighted average of models rather than selecting a single model.
\subsection{Synthetic Dataset} 
This section presents a simulation study on a regression task to evaluate our proposed methods. The data generating process is defined as:
\begin{align} \label{eq:dgp1}
    y_i = \beta_0 +\beta_{1}x_{i1}+\beta_{2}x_{i2}+\alpha_1x_{i1}^2  +\alpha_2x_{i2}^2 + \gamma x_{i1}x_{i2} + \sin^2(x_{i1}) + \epsilon_i, 
\end{align}
where observed predictors $\bx_i = (x_{i1},x_{i2})^\top$ are independent identically distributed (i.i.d.) from $\mathcal{N}(\mathbf{0},\mathbf{I})$, and $\epsilon_i \sim \mathcal{N}(0,\sigma^2)$. 
The coefficients \( (\beta_0,\beta_1, \beta_2, \alpha_1, \alpha_2, \gamma) \) are drawn independently from \( \mathcal{N}(0, 0.09) \). We vary noise levels with \( \sigma \in \{0.1,0.5,0.9,1.5 \} \) and labeled sample sizes \( n \in \{100, 200, 400, 800\} \) for labeled data, fixing unlabeled sample size \( N = 2000 \). Each configuration is repeated 100 times.

\begin{figure}[htbp]
    \centering
    \includegraphics[width=\linewidth]{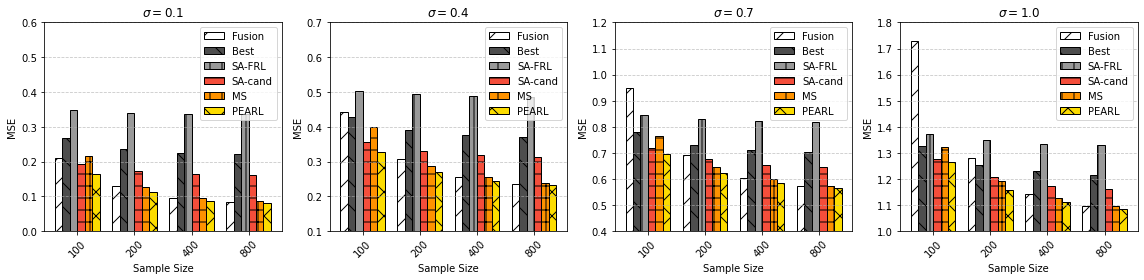}
    \caption{Synthetic-data results}
    \label{fig:sim}
\end{figure}

We adopt kernel principal component analysis (KPCA) \citep{mika1998kernel} with five kernels shown in Table \ref{tab:methods_summary} as FRL models to extract representations from predictors. These yield five foundation representation sets, from which we construct \( J = 2^5 - 1 = 31 \) candidate sets by all non-empty combinations (see Example~\ref{example: candidate}). For each candidate set, a linear regression is fit on the labeled dataset to predict the target variable.

The results in Fig.~\ref{fig:sim} reveal that our method outperforms the baseline methods in most cases. As the noise level increases, which is proportional to $\sigma$, the performance of the Fusion and MS methods deteriorates more rapidly than that of PEARL, particularly when \( n = 100 \) or \( 200 \). For larger sample sizes (\( n = 500 \) or \( 1000 \)), PEARL substantially outperforms Best, SA-FRL and SA-cand. These results demonstrate that PEARL mitigates overfitting in small sample sizes and maximizes predictive performance in larger datasets.

\subsection{MNIST Dataset} \label{sec:mnist}
We also evaluate PEARL on the MNIST benchmark for image classification. We use similar configuration for FRL methods, except that the ResNet encoders are replaced with CNN ones. For downstream classification, we employ either a simple MLP or multinomial logistic regression. As shown in Table~\ref{tab_exp:5}, PEARL consistently outperforms baselines in MNIST dataset. The implementation details are provided in Appendix B.

We evaluate PEARL on the MNIST benchmark for image classification, consisting of 70k grayscale digit images (60k training, 10k testing,  $28\times28$ pixels). We use two types of FRL methods, AE and SimCLR \citep{chen2020simple}, each with MLP and CNN encoders, and also include the raw images as an additional representation set. Therefore, we have $M=5$ FRL and $J=31$ candidate representation sets via Example~\ref{example: candidate}. For downstream classification, we employ either a simple MLP or multinomial logistic regression. Experiments are conducted with \( n \in \{500, 1000, 3000, 6000\} \) labeled samples, repeated 50 times with random subsets, while FRL models are pretrained once on the full training set to enable reuse across tasks.

For evaluation, we use the standard 10000-image test set and measure the performance of different methods using three metrics: classification accuracy (Acc), CE, and mean squared error (MSE) between the true and predicted labels. In Table~\ref{tab_exp:5} and all subsequent experiments, the best results are highlighted in bold red, while the second-best are underlined blue.

Table~\ref{tab_exp:5} demonstrates that PEARL consistently outperforms baseline methods across all metrics and sample sizes, indicating superior classification performance and enhanced calibration. The implementation details are provided in Appendix B.

\begin{table}[ht]
\centering
\caption{Results of different methods on MNIST dataset}
\label{tab_exp:5}
\resizebox{\linewidth}{!}{ 
\begin{tabular}{@{}llcccccc@{}}
\toprule
 & \textbf{n} & \textbf{Fusion} & \textbf{Best} & \textbf{SA-FRL} & \textbf{SA-cand} & \textbf{MS} & \textbf{PEARL} \\
\midrule
\multirow{4}{*}{Acc} & 300  & 94.647 (0.509) & 95.390 (0.385) & 94.140 (0.501) & 94.863 (0.467) & \secondres{96.242 (0.478)} & \boldres{96.355 (0.451)} \\
 & 600  & 95.949 (0.309) & 96.016 (0.236) & 95.064 (0.295) & 96.062 (0.282) & \secondres{96.984 (0.286)} & \boldres{97.086 (0.237)} \\
 & 1500  & 97.066 (0.204) & 96.879 (0.165) & 96.095 (0.191) & 97.241 (0.193) & \secondres{97.625 (0.187)} & \boldres{97.762 (0.159)} \\
 & 3000  & 97.570 (0.143) & 97.316 (0.145) & 96.578 (0.133) & 97.844 (0.120) & \secondres{97.939 (0.165)} & \boldres{98.090 (0.122)} \\
\midrule
\multirow{4}{*}{CE} & 300  & 0.170 (0.016) & 0.152 (0.015) & 0.238 (0.011) & 0.178 (0.011) & \secondres{0.122 (0.015)} & \boldres{0.121 (0.013)} \\
 & 600  & 0.132 (0.011) & 0.127 (0.007) & 0.196 (0.006) & 0.140 (0.006) & \secondres{0.096 (0.009)} & \boldres{0.094 (0.006)} \\
 & 1500  & 0.099 (0.008) & 0.099 (0.005) & 0.161 (0.003) & 0.106 (0.004) & \secondres{0.074 (0.006)} & \boldres{0.071 (0.004)} \\
 & 3000  & 0.082 (0.006) & 0.085 (0.003) & 0.146 (0.002) & 0.090 (0.002) & \secondres{0.064 (0.005)} & \boldres{0.059 (0.003)} \\
\midrule
\multirow{4}{*}{MSE} & 300  & 0.008 (0.001) & 0.007 (0.001) & 0.011 (0.001) & 0.008 (0.001) & \secondres{0.006 (0.001)} & \boldres{0.006 (0.001)} \\
 & 600  & 0.006 (0.000) & 0.006 (0.000) & 0.009 (0.000) & 0.006 (0.000) & \secondres{0.005 (0.000)} & \boldres{0.004 (0.000)} \\
 & 1500  & 0.004 (0.000) & 0.005 (0.000) & 0.007 (0.000) & 0.005 (0.000) & \secondres{0.004 (0.000)} & \boldres{0.003 (0.000)} \\
 & 3000  & 0.004 (0.000) & 0.004 (0.000) & 0.007 (0.000) & 0.004 (0.000) & \secondres{0.003 (0.000)} & \boldres{0.003 (0.000)} \\
\bottomrule
\end{tabular}
}
\end{table}

\subsection{CIFAR-10 Dataset}
The CIFAR-10 dataset  contains 60,000 color images of size 32×32 pixels, divided into 10 object categories (airplane, automobile, bird, cat, deer, dog, frog, horse, ship, and truck). The dataset is split into 50,000 training images and 10,000 test images, with 6,000 images per class. We use $M=4$ FRL models as shown in Table \ref{tab:methods_summary}, which create $J=15$ candidate representation sets via Example~\ref{example: candidate}. For AE, VAE, and SimCLR we use ResNet encoders, where SimCLR is also instantiated with a CNN encoder. See Appendix D for implementation details. For downstream classification, we employ either a simple MLP or multinomial logistic regression. Experiments are conducted with \( n \in \{300, 500, 1500, 3000\} \) labeled samples, repeated 100 times with random subsets, while FRL models are pretrained once on the full training set to enable reuse across tasks.
For evaluation, we use three metrics: classification accuracy (Acc), CE, and mean squared error (MSE) between the true and predicted labels/probabilities. 

As shown in Table~\ref{tab:cifar}, PEARL consistently outperforms baselines across all metrics and sample sizes, demonstrating superior classification accuracy and better calibration. Compared with MS, which successfully selects the single FRL model with the highest performance since MS has the same performance with Best, PEARL further improves the results. This demonstrates the clear advantage of combining different models rather than relying on a single one. Moreover, when compared with SA-FRL and SA-cand, PEARL also achieves superior performance. This highlights that different models contribute unequally to the downstream task, and assigning task-specific weights is more effective than applying equal weights. Finally, relative to Fusion, the weighted averaging strategy in PEARL yields better results. This confirms that learning to weight different representation groups is more beneficial than simply concatenating them.

 \begin{table}[ht]
\centering
\caption{Results of different methods on CIFAR-10 dataset}
\label{tab:cifar}
\resizebox{\linewidth}{!}{ 
\begin{tabular}{@{}llcccccc@{}}
\toprule
 & \textbf{n} & \textbf{Fusion} & \textbf{Best} & \textbf{SA-FRL} & \textbf{SA-cand} & \textbf{MS} & \textbf{PEARL} \\
\midrule
\multirow{4}{*}{Acc (\%)} & 300  & 42.129 (1.938) & \secondres{53.379 (0.836)} & 42.125 (2.441) & 43.839 (1.992) & \secondres{53.379 (0.836)} & \boldres{53.825 (0.831)} \\
 & 600  & 48.748 (1.600) & \secondres{56.133 (0.529)} & 48.449 (1.416) & 50.791 (1.527) & \secondres{56.133 (0.529)} & \boldres{57.028 (0.521)} \\
 & 1500  & 56.389 (0.706) & \secondres{58.432 (0.380)} & 53.726 (0.775) & 57.663 (0.701) & \secondres{58.432 (0.380)} & \boldres{60.186 (0.375)} \\
 & 3000  & 60.342 (0.456) & 59.518 (0.251) & 55.791 (0.493) & \secondres{60.704 (0.456)} & 59.817 (0.566) & \boldres{61.928 (0.356)} \\
\midrule
\multirow{4}{*}{CE} & 300  & 2.589 (0.303) & \secondres{1.292 (0.023)} & 1.587 (0.031) & 1.614 (0.075) & \secondres{1.292 (0.023)} & \boldres{1.282 (0.023)} \\
 & 600  & 1.983 (0.220) & \secondres{1.212 (0.011)} & 1.475 (0.013) & 1.399 (0.041) & \secondres{1.212 (0.011)} & \boldres{1.190 (0.012)} \\
 & 1500  & 1.350 (0.047) & \secondres{1.149 (0.006)} & 1.404 (0.006) & 1.233 (0.013) & \secondres{1.149 (0.006)} & \boldres{1.104 (0.007)} \\
 & 3000  & 1.140 (0.015) & 1.120 (0.004) & 1.378 (0.005) & 1.173 (0.007) & \secondres{1.114 (0.010)} & \boldres{1.054 (0.006)} \\
\midrule
\multirow{4}{*}{MSE} & 300  & 0.087 (0.005) & \secondres{0.060 (0.001)} & 0.071 (0.002) & 0.072 (0.003) & \secondres{0.060 (0.001)} & \boldres{0.059 (0.001)} \\
 & 600  & 0.075 (0.004) & \secondres{0.057 (0.000)} & 0.067 (0.001) & 0.064 (0.002) & \secondres{0.057 (0.000)} & \boldres{0.056 (0.000)} \\
 & 1500  & 0.060 (0.001) & \secondres{0.054 (0.000)} & 0.064 (0.000) & 0.057 (0.001) & \secondres{0.054 (0.000)} & \boldres{0.052 (0.000)} \\
 & 3000  & 0.053 (0.001) & 0.053 (0.000) & 0.063 (0.000) & 0.054 (0.000) & \secondres{0.053 (0.001)} & \boldres{0.050 (0.000)} \\
\bottomrule
\end{tabular}
}
\end{table}

\subsection{Polarity Dataset} \label{sec:polarity}
We use the Polarity v2.0 dataset \citep{PangLee04a}, a standard benchmark for sentiment analysis containing 2000 movie reviews evenly split between positive and negative labels. The task is binary sentiment classification. We consider $M=6$ FRL models, including CountVectorizer, TF-IDF (term frequency-inverse document frequency), Doc2Vec \citep{le2014distributed}, Word2Vec \citep{mikolov2013efficient}, BERT, and GloVe \citep{pennington2014glove}. Among these, BERT and GloVe are pre-trained models, while the others are trained from scratch. For downstream classification, we apply either a simple MLP or logistic regression on the learned representations, using a 70\%-30\% train–test split repeated 10 times for robustness.

\begin{table}[ht]
\centering
\caption{Comparison of different methods on polarity dataset}
\label{tab_exp:2}
\begin{tabular}{@{}lcccc@{}}
\toprule
 & \textbf{Fusion} & \textbf{Best} & \textbf{SA} & \textbf{PEARL-CE} \\
\midrule
Acc (\%) & 83.587 (1.445) & \secondres{87.745 (1.113)} & 86.233 (1.392) & \boldres{88.873 (1.173)} \\
\midrule
CE & 0.482 (0.045) & \secondres{0.308 (0.014)} & 0.423 (0.009) & \boldres{0.291 (0.013)} \\
\midrule
MSE & 0.126 (0.010) & \secondres{0.092 (0.005)} & 0.128 (0.004) & \boldres{0.086 (0.005)} \\
\bottomrule
\end{tabular}
\end{table}


Given the high dimensionality of CountVectorizer and TF-IDF, no additional candidate sets are constructed, yielding \( M = J = 6 \).  In this case, SA-cand reduces to SA-FRL, and MS is dominated by Best, so both are omitted from comparison. As shown in Table~\ref{tab_exp:2}, PEARL consistently outperforms the remaining baselines, demonstrating its effectiveness even without additional candidate sets.

\subsection{Wine Review Dataset} \label{sec:wine_review}
We evaluate PEARL in a multi-modal setting using the Wine Review dataset (approximately 130k records), where each review contains unstructured text and structured attributes (variety, location, winery, and price). The task is to predict the wine quality score as a continuous outcome. For representation learning, we use PCA and VAEs \citep{kingma2013auto} for numerical attributes, and pre-trained BERT and GPT-2 \citep{radford2019language} for text, trained on 10k unlabeled samples. Linear regression is applied to labeled sets of size \( n \in \{100, 200, 500, 1000, 2000\} \), with evaluation on 5k test samples. Each split is repeated 10 times, and performance is assessed by average MSE.


\begin{table}[ht]
\centering
\caption{Comparison of different methods on wine review dataset}
\label{tab:wrd}
\resizebox{\linewidth}{!}{ 
\begin{tabular}{@{}rcccccc@{}}
\toprule
\textbf{n} & \textbf{Best} & \textbf{Fusion} & \textbf{SA-repr} & \textbf{SA-cand} & \textbf{MS} & \textbf{PEARL} \\
\midrule
100 & 7.296 (0.588) & \secondres{6.159 (0.537)} & 26.849 (56.176) & 17.670 (28.147) & 6.468 (0.592) & \boldres{5.726 (0.384)} \\
200 & 7.596 (0.451) & 5.678 (0.353) & 5.251 (0.291) & \boldres{4.973 (0.274)} & 6.203 (0.792) & \secondres{5.103 (0.334)} \\
500 & 7.124 (0.335) & 6.497 (0.350) & 5.441 (0.264) & \secondres{5.371 (0.268)} & 6.557 (0.393) & \boldres{4.920 (0.193)} \\
1000 & 6.730 (0.264) & 10.336 (0.659) & \secondres{5.407 (0.206)} & 6.171 (0.286) & 6.878 (0.550) & \boldres{5.332 (0.247)} \\
2000 & 5.496 (0.136) & 14.210 (1.002) & \secondres{4.147 (0.131)} & 4.416 (0.127) & 5.043 (0.299) & \boldres{3.659 (0.084)} \\
\bottomrule
\end{tabular}
}
\end{table}

From Table~\ref{tab:wrd}, across most sample sizes, PEARL achieves the lowest MSE compared with both single-modality baselines (Best) and other averaging strategies, showing clear advantages in integrating heterogeneous representations from multi-modal datasets. These results underscore that PEARL not only improves prediction accuracy but also maintains robustness in the challenging multi-modal setting, highlighting its strong generalizability beyond unimodal tasks.
 
\section{Summary}\label{Summary}
In this work, we introduce a theoretically grounded aggregation method for representation learning based on frequentist model averaging. The proposed framework is widely applicable and highly general, accommodating tasks ranging from matrix/tensor decomposition to nonlinear representation learning in deep neural networks. This broad scope underscores its ability to bridge interpretable statistical tools with modern AI techniques, making it suitable for diverse real-world applications.

Future work can focus on developing online learning algorithms that enable the model ensemble to adapt in real-time to evolving data streams, ensuring it remains responsive and robust in non-stationary environments. Additionally, it is appealing to applying the framework to deep learning in decision-making applications in areas such as autonomous driving, medical diagnostics, or financial risk assessment. 

\bibliographystyle{chicago}
\bibliography{reference}

	
	
		
		
	

\end{document}